\pdfoutput=1

\documentclass[11pt]{article}

\usepackage[final]{acl}

\usepackage{times}
\usepackage{latexsym}
\usepackage{graphicx}
\usepackage{subfigure}
\usepackage{booktabs}
\usepackage[T1]{fontenc}

\usepackage[utf8]{inputenc}

\usepackage{microtype}

\usepackage{inconsolata}

\title{MS4UI: A Dataset for Multi-modal Summarization  \\of User Interface Instructional Videos}

\author{ 
 \textbf{Yuan Zang\textsuperscript{1}\thanks{Work done while the first author was an intern at Adobe Research}},
 \textbf{Hao Tan\textsuperscript{2}},
 \textbf{Seunghyun Yoon\textsuperscript{2}},
 \textbf{Franck Dernoncourt\textsuperscript{2}},\\
 \textbf{Jiuxiang Gu\textsuperscript{2}},
 \textbf{Kushal Kafle\textsuperscript{2}},
 \textbf{Chen Sun\textsuperscript{1}},
 \textbf{Trung Bui\textsuperscript{2}}
\\
 \textsuperscript{1}Brown University
 \textsuperscript{2}Adobe Research
\\
 \small{
    \href{mailto:yuan_zang@brown.edu}{yuan\_zang@brown.edu}, \href{mailto:bui@adobe.com}{bui@adobe.com}
 }
}

\begin{document}
\maketitle
\begin{abstract}
We study multi-modal summarization for instructional videos, whose goal is to provide users an efficient way to learn skills in the form of text instructions and key video frames.
We observe that existing benchmarks focus on generic semantic-level video summarization, and are not suitable for providing step-by-step executable instructions and illustrations, both of which are crucial for instructional videos.
We propose a novel benchmark for user interface (UI) instructional video summarization to fill the gap. %
We collect a dataset of 2,413 UI instructional videos, which spans over 167 hours. These videos are manually annotated for video segmentation, text summarization, and video summarization, which enable the comprehensive evaluations for concise and executable %
video summarization.
We conduct extensive experiments on our collected MS4UI dataset, which suggest that state-of-the-art multi-modal summarization methods struggle on UI video summarization, and highlight the importance of new methods for UI instructional video summarization.
\end{abstract}
\section{Introduction}
\label{sec:intro}

Multi-modal summarization proposes an effective way to learn from lengthy instructional videos. Researchers have proposed datasets and methods for multi-modal summarization of news \citep{li-etal-2020-vmsmo,zhu-etal-2018-msmo}, movies \citep{rao2020local} and daily tasks \citep{sanabria2018how2}. Meanwhile, data and methods for multimodal summarization of User Interface (UI) instructional videos are scarce. Compared with daily task videos, it is more challenging to comprehend and summarize UI instructional videos. First, the modeling of UI and UI-related videos requires deep understanding of the visual elements in the UIs, which are structured, symbolic and abstract. Furthermore, in order to align the technical terms in the text instructions and the visual elements in the video, it requires domain-specific and fine-grained vision-language grounding. Recently, many UI related tasks including command language grounding \citep{li2020mapping}, UI element retrieval \citep{he2021actionbert}, UI description generation \citep{huang2019swire} have been proposed. In this paper, we focus on summarizing UI instructional videos into compact and executable step-by-step instructions.
\begin{figure}[!t]
\centering
\vspace{-2em}
\includegraphics[width=\linewidth]{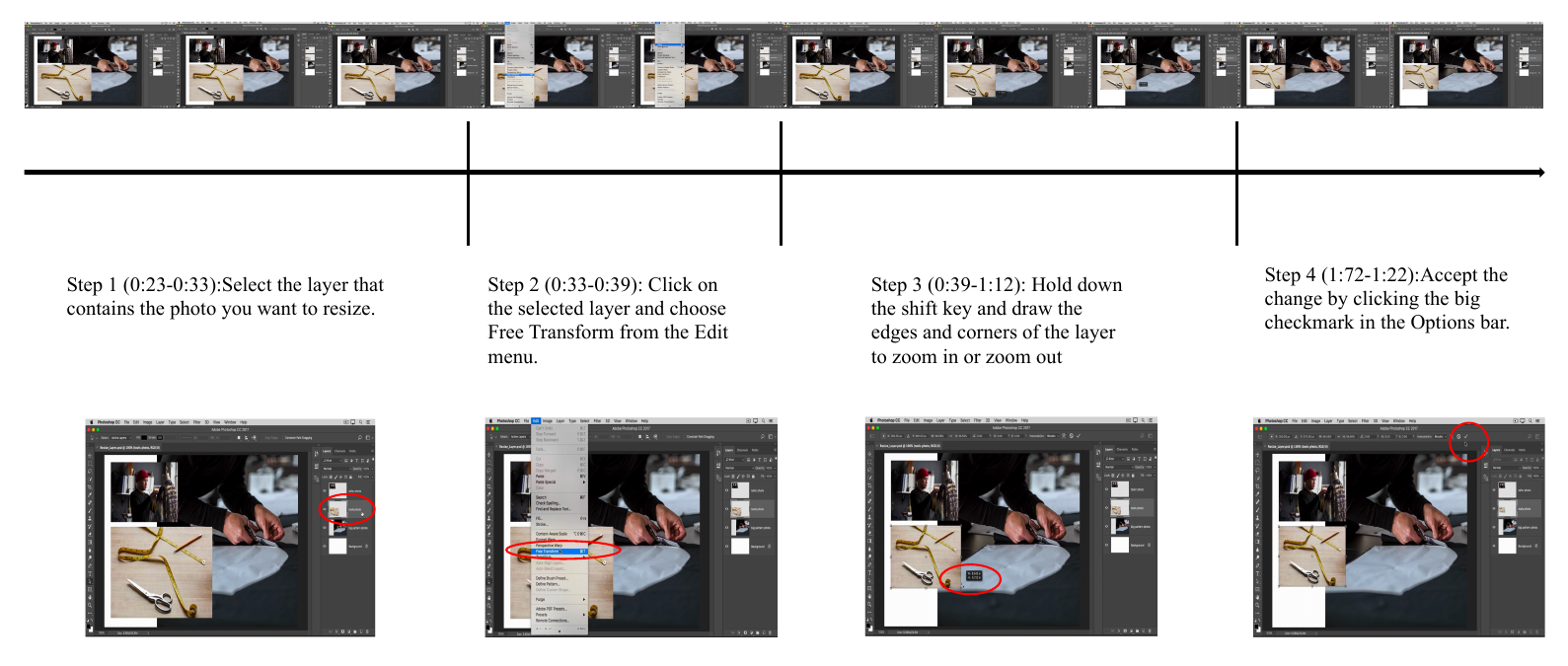}
\caption{An example of the summarization of UI tutorial videos in our dataset. It shows the summarization of the task ``Resize the layer in Adobe Photoshop''. First, we segment the video into different steps. For each step, we summarize the key operations into concise text instruction and select a key image frame that describing the corresponding operation. The red circles note the key elements related to the operations in the instruction.}
\label{fig:example}
\vspace{-0.4cm}
\end{figure}

Solving the proposed tasks of UI instructional video summarization is challenging. Existing multi-modal summarization methods and benchmarks, mainly focus on document-level summarization (e.g. a topic sentence and cover image for the news video), are not suitable for UI instructional video summarization because of following reasons. First, the instructional videos usually contain steps and thus require step-by-step summarization, but most existing multi-modal summarization methods do not support step segmentation. In addition, while existing multi-modal summarization tasks mainly stress semantic-remaining summarization, the step summarization for UI instructional videos should be executable and contain accurate technical details for specific actions. For example, to be executable, the instruction for a certain action usually contain the details like which item to select in the toolbar and how to set the parameters, which might be neglected in general-purpose summarization. Therefore, we propose three core tasks for UI instructional video summarization, \textbf{Video Segmentation}, \textbf{Text Summarization} and \textbf{Video Summarization}. Given an instructional video, we first segment it into different steps. Then we generate the text instruction for each step according to the video content. To make the instructions easy-to-execute, we select representing frame which demonstrates the key actions for each step from the video. Figure \ref{fig:example} shows an example the three tasks.

In this paper, we collect a multi-modal summarization dataset of 2,413 UI instructional videos for Adobe Creative Cloud UIs from media including Adobe HelpX and Youtube and hire human annotators to perform the proposed tasks on the collected videos. We develop evaluation metrics for the proposed tasks and perform comprehensive experiments to evaluate state-of-the-art video summarization methods on the dataset. The results indicate that existing methods struggle with understanding and summarizing UI videos in the proposed dataset.

In short, we make three key contributions:

\begin{itemize}
    \item We propose a novel multi-modal summarization dataset, which firstly focus on UI instructional video summarization.
    \item We introduce three core tasks and develop comprehensive evaluation metrics for the purpose of generating concise and executable step-by-step multi-modal summarization.
    \item We conduct exhaustive experiments to evaluate existing state-of-the-art methods on our dataset and demonstrate the significance of paying attention to the unique features of UI instructional video summarization.
\end{itemize}

\section{Related Work}
\label{sec:related}
With the development of multi-modal learning and its application on multi-media contents like videos, multi-modal summarization has attracted more and more attention. Unlike traditional video summarization \citep{gygli2015video,zhang2016summary} and text summarization \citep{nallapati2016neural,zhu2018augmenting,celikyilmaz2018deep} that only rely on single vision or text modality, researchers \citep{krubinski2023mlask,he2023align,fu2020multi} have proposed to utilize multi-modal fusion modules to introduce the information from different modality into the representation of videos and texts to generate vision and text summarization. Researchers have developed benchmarks for multi-modal summarization of videos in various domains, including news \citep{zhu-etal-2018-msmo,li-etal-2020-vmsmo}, medical analysis \citep{liu2022machine}, movies \citep{rao2020local} and daily tasks \citep{song2015tvsum,gygli2014creating,sanabria2018how2}. However, most of those benchmarks focus on news or story video summarization, and the summarization of instructional videos is scarce.

Previous work proposed to identify the significant frames in the instructional videos \citep{narasimhan2022tl} to build the summarization. However, it lacks clear step segmentation and textual instruction. \citet{sanabria2018how2} collected instructional video data to build dataset How2, which contains the textual instruction for each step of the instructional videos but lacks the key image frames.
In this paper, we build a multi-modal summarization dataset containing segmentation as well as text and video summarization for UI instructional videos.

\section{MS4UI Dataset}
\label{sec:dataset}

\subsection{Data Collection}
We propose a new dataset for multi-modal summarization of UI instructional videos. We collect instructional videos for Adobe Creative Cloud products from Adobe Support \footnote{https://helpx.adobe.com/support/creative-cloud.html} and Youtube \footnote{https://www.youtube.com}. The collected videos include tutorials for various UIs including Photoshop, Illustrator, Acrobat and Premiere. These tutorials provide detailed illustration of diverse functions in these UIs, such as editing images, modifying PDFs, and editing videos.  
\subsection{Multi-modal Summarization and Human Verification}
In order to generate step-by-step summarization for the tutorial videos, we utilize GPT-3.5 to segment and summarize the transcriptions of these videos. Utilizing pre-designed prompts and examples, we can obtain step segmentation with time stamps and text summarization for each step that includes key operations. We ask human workers to verify the generated segmentation and summarization to ensure the fidelity. The human workers are tasked with revising the step segmentation, which includes adding necessary steps and removing duplicate ones, as well as verifying the time interval of each step and revising the step summarization. For the video summarization task, we ask human workers to select representing image frame that describes key operations for each step. 
\subsection{Dataset Statistics}
\begin{figure}
  \centering
  \begin{subfigure}{}
    \includegraphics[width=0.2\textwidth]{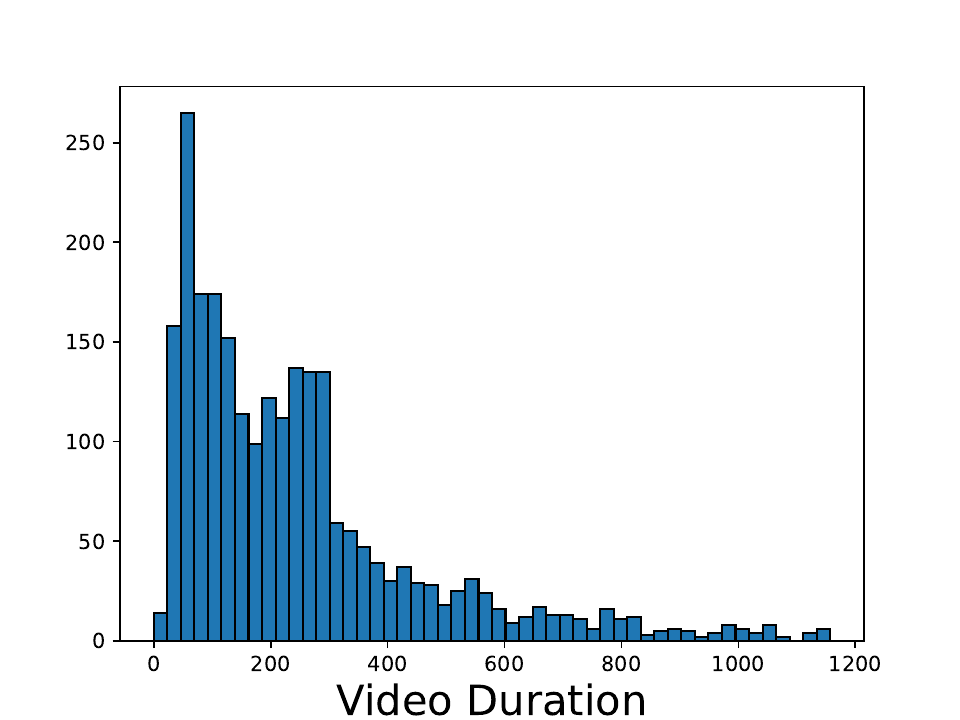}
    \label{fig:sub1}
  \end{subfigure}
  \begin{subfigure}{}
    \includegraphics[width=0.2\textwidth]{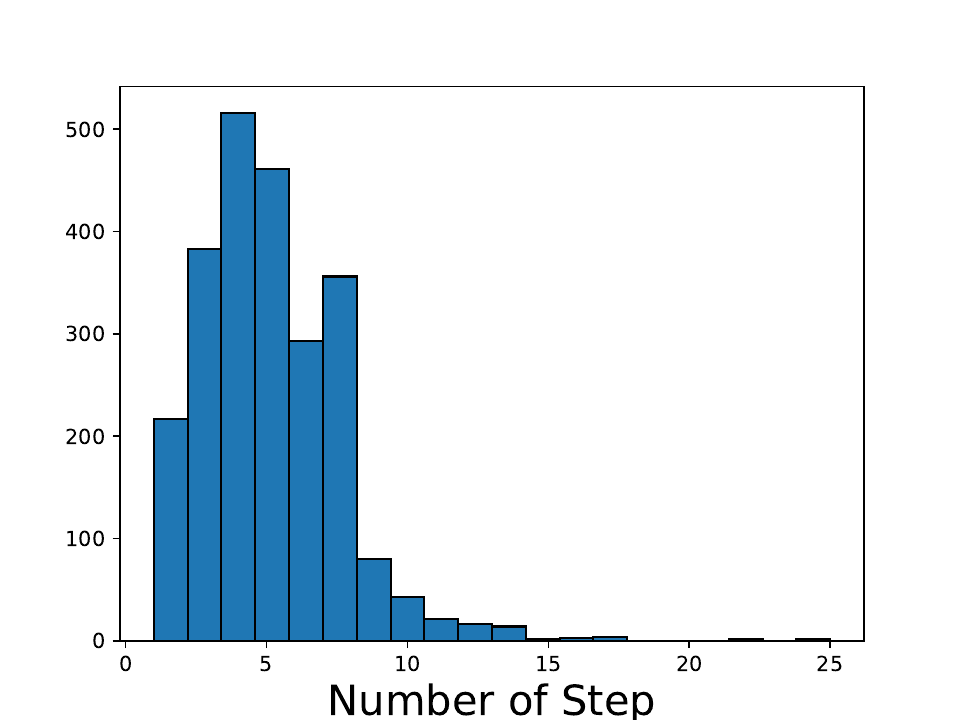}
    \label{fig:sub2}
  \end{subfigure}

  \begin{subfigure}{}
    \includegraphics[width=0.2\textwidth]{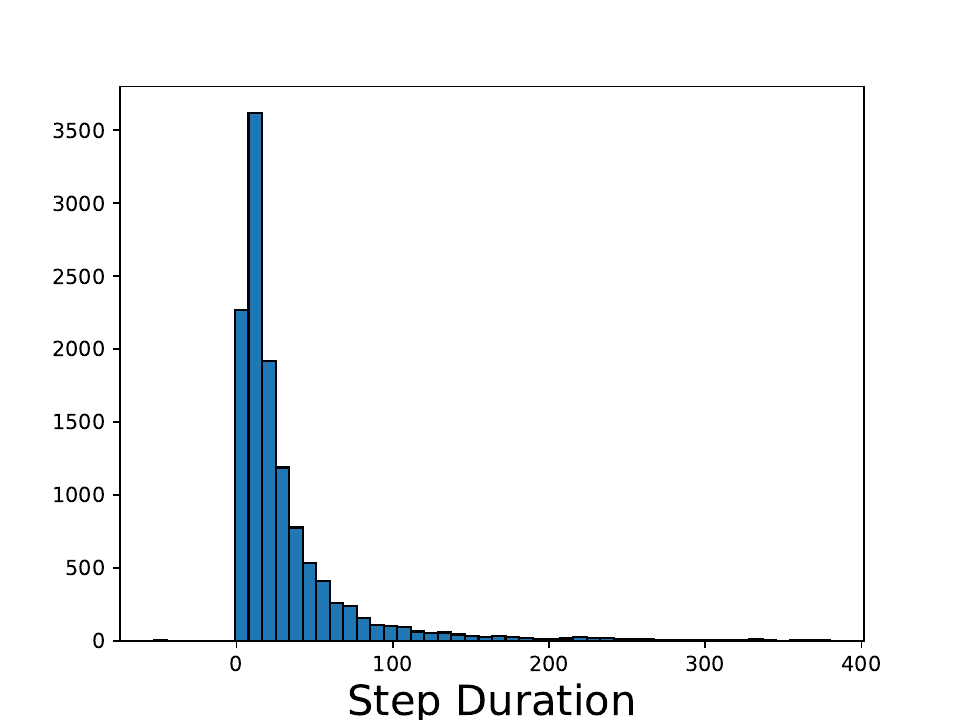}
    \label{fig:sub3}
  \end{subfigure}
  \begin{subfigure}{}
    \includegraphics[width=0.2\textwidth]{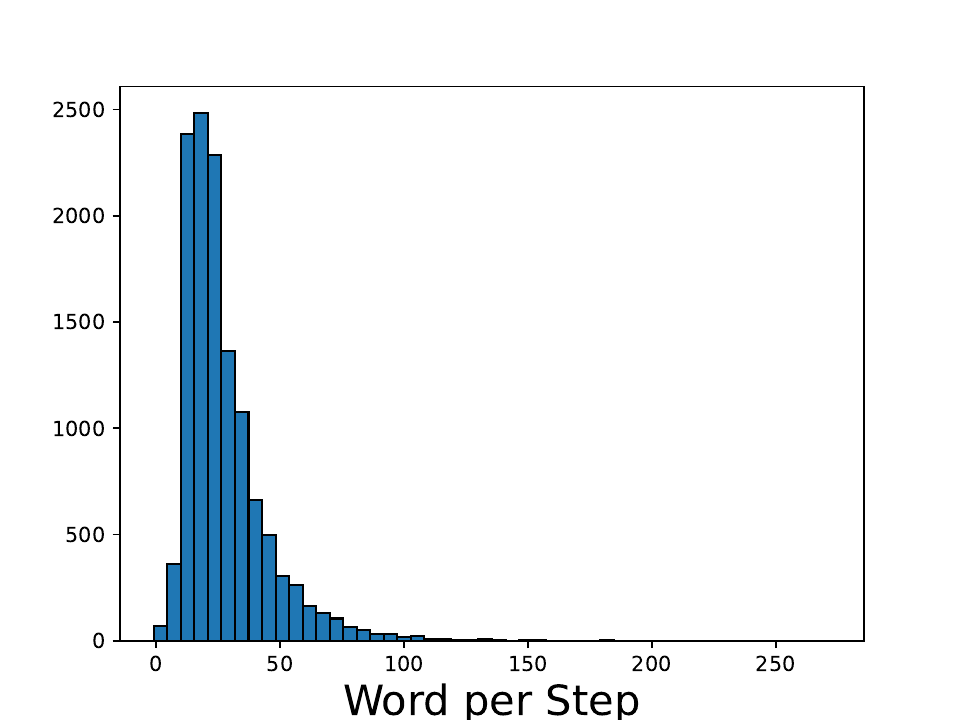}
    \label{fig:sub4}
  \end{subfigure}
  \caption{Detailed statistics of the dataset, which show the distribution of the video duration, step numbers, step duration and step summarization length.}
  \label{fig:statistics}
  \vspace{-0.4cm}
\end{figure}

Our dataset contains 2,413 UI instructional videos of 167 hours length in total. Each video contains 5.08 steps in average. Each step is 32.47 second long and contains 27.28 words in average. Figure \ref{fig:statistics} shows the detailed statistics of the proposed dataset.
\begin{table}[!t]
\centering
\resizebox*{\linewidth}{!}{
\begin{tabular}{l|r|r|r|r|r}
\toprule
\multicolumn{1}{l}{}       & \multicolumn{1}{c}{TVSum}          & \multicolumn{1}{c}{SumMe}          & \multicolumn{1}{c}{VMSMO} & \multicolumn{1}{c}{How2}                 & \multicolumn{1}{c}{MS4UI}                   \\ \midrule
Type                       & \multicolumn{1}{c}{General} & \multicolumn{1}{c}{General} & \multicolumn{1}{c}{News}  & \multicolumn{1}{c}{Instruction} & \multicolumn{1}{c}{UI Instruction} \\ \midrule
Num of Video               & 50                                 & 25                                 & 184,920                   & 79,114                                   & 2,413                                       \\\midrule
Total Duration (Hours)     & 4                                  & 1                                  & 3082                      & 2000                                     & 167                                         \\\midrule
Avg. Video Duration (Mins) & 3.9                                & 2.4                                & 1.0                       & 1.5                                      & 4.2                                         \\\midrule
Have Segmentation          & No                                 & No                                 & No                        & Yes                                      & Yes                                         \\\midrule
Avg. Num of Key Frame      & 70                                 & 44                                 & 1                         & NA                                       & 5                                           \\\midrule
Avg. Text Summary Length   & NA                                 & NA                                 & 11                        & 20                                       & 27                                 \\\bottomrule        
\end{tabular}
}
\caption{Comparison of different dataset for video summarization.}
\label{tab:compare}
\end{table}
\subsection{Comparison with Existing Datasets}
Table \ref{tab:compare} provides a comparison between the proposed dataset and existing video summarization datasets.
Previous datasets such as TVSum \citep{song2015tvsum} and SumMe \citep{gygli2014creating} primarily emphasize visual aspects and do not include segmentation or text summaries. While news summarization datasets like VMSMO \citep{li-etal-2020-vmsmo} do include text summaries, they often lack segmentation due to the short duration of news videos compared to instructional videos. The How2 dataset focuses on instructional videos and provides segmentation and text summaries, but it does not include key frame annotations that illustrate the actions in the instructions. The proposed dataset is the first dataset containing segmentation and multi-modal summarization tasks.

\section{Experiments}
\label{sec:benchmark}
\subsection{Baseline Models}
\paragraph{Video Segmentation}
For video segmentation, we compare text-based and vision based baselines. For text-based segmentation, we implement Cross Textseg \citep{lukasik-etal-2020-text} which utilizes a hierarchy Transformer architecture to firstly encode the sentences and then identify the boundary sentences based on the sentence embeddings. For vision-based segmentation, we implement LGSS \citep{rao2020local} which proposes a boundary network to model the shot boundary for video segmentation. We also utilize the recently released video segmentation toolbox PySceneDetect \citep{castellano2021intelligent} as a vision-based baseline. 
\vspace{-0.2cm}
\paragraph{Text Summarization}
For text summarization, we firstly utilize pre-trained language models including BERT2BERT \citep{chen2021bert2bert}, BART \citep{lewis2019bart} and T5 \citep{raffel2020exploring} to summarize the transcriptions for each step into concise instructions. In order to utilize the vision information, we also implement multi-modal summarization frameworks MLASK \citep{krubinski2023mlask} and A2Summ \citep{he2023align}.
\vspace{-0.2cm}
\paragraph{Video Summarization}
For video summarization, we also use the multi-modal summarization models MLASK and A2Summ to model the image frames and transcriptions simultaneously and select the key image frames. We also compare vision-only method VSumm \citep{de2011vsumm} which extracts image features with VGG16 \citep{simonyan2014very} and utilizes K-means to cluster the features to select the key frame. 
\subsection{Evaluation Metrics}
\paragraph{Video Segmentation} 
We introduce Intersection over Union (IOU) to evaluate different video segmentation methods. Considering segmentations $g$ and $p$ as unions of image frames, the IOU is calculated as
\begin{equation}
    \mathrm{IOU} = \frac{g\cap p}{g\cup p}
\end{equation}
Given the ground-truth video segmentation $G=\{g_1, g_2, ..., g_n\}$ and the predicted segmentation $P=\{p_1, p_2, ..., p_m\}$, we can calculate the mean IOU (MIOU) for the predicted segmentations as 
\begin{equation}
   \mathrm{MIOU} = \frac{1}{m}\sum_{i=0}^{m}\max_{j}\mathrm{IOU}(p_i,g_j).
\end{equation}
The MIOU metric is widely used to evaluate the segmentation accuracy in the domain of video. Based on IOU, we can also calculate the Precision, Recall and F1 score of the predicted segmentations. Given an IOU threshold $th$, the predicted segmentation that has an IOU high than $th$ with a ground-truth segmentation is regarded as a true positive prediction. Following previous works \citep{rao2020local,qiu2023multisum}, we report the MIOU and F1 with different IOU threshold (0.1, 0.25, 0.5) for the evaluation of video segmentation.
\paragraph{Text Summarization}
For text summarization, we introduce the Recall-Oriented Understudy for Gisting Evaluation (ROUGE) \citep{lin-2004-rouge} as the evaluation metric. The ROUGE measures the overlap between the n-grams of the generated summarization and the reference summarization verified by human. Following previous works, we utilize the ROUGE-1, ROUGE-2 and ROUGE-2 metrics to evaluate the text summarization methods.
\begin{table}[!t]

\resizebox*{\linewidth}{!}{
\begin{tabular}{l|r|r|r|r}
\toprule
              & MIOU & F1@0.1 & F1@0.25 & F1@0.5 \\ \midrule
Random & 14.61 & 34.94 & 25.20& 8.62\\ \midrule
Cross TextSeg & \textbf{20.53} & 43.47   & \textbf{31.24}    & \textbf{12.77}   \\ \midrule
LGSS          & 20.08 &  \textbf{44.04}  & 30.58    & 11.63   \\ \midrule
PySceneDetect & 19.21 & 41.55   & 29.89    & 11.73 \\ \bottomrule
\end{tabular}
}

\caption{Video segmentation results of baseline methods.} %
\label{tab:seg}

\end{table}

\begin{table}[!t]
\centering
\resizebox*{0.8\linewidth}{0.15\textheight}{
\begin{tabular}{l|r|r|r}
\toprule
          & Rouge-1 & Rouge-2 & Rouge-L \\ \midrule
Random    & 2.77        &  0.31       &    2.56     \\ \midrule
BERT2BERT & 2.78    & 0.28    & 2.53    \\\midrule
T-5       & 2.89    & 0.47    & 2.64    \\\midrule
BART-CNN  & 3.12    & 0.44    & 3.01    \\\midrule
BART-XSUM & 3.23    & 0.42    & 3.10    \\\midrule
MLASK     & \textbf{5.10}    & \textbf{1.31}    & \textbf{3.98}    \\\midrule
A2Summ    & 3.82    & 0.65    & 3.27   \\ \bottomrule
\end{tabular}
}
\caption{Text summarization results of baseline methods.}
\label{tab:textsum}
\vspace{-0.5cm}
\end{table}

\begin{table}[!t]
\centering
\resizebox*{0.8\linewidth}{0.1\textheight}{
\begin{tabular}{l|r|r|r}
\toprule
      & Recall @ 1 & Recall @ 2 & Recall @ 5 \\ \midrule
Random &   3.09       &    6.18        &    15.62        \\ \midrule
Vsumm & 7.78       & 11.74      & 22.58      \\ \midrule
MLASK & 10.46      & 14.35      & 27.47      \\ \midrule
ASumm & 9.82       & 13.20      & 27.21     \\ \bottomrule
\end{tabular}
}
\caption{Video summarization results of baseline methods.}
\label{tab:videosum}
\vspace{-0.5cm}
\end{table}

\paragraph{Video Summarization}
For video summarization, we select image frames from the video with regular intervals (1fps) and regard the image closest to the human-labeled key image frame as the ground-truth frame. We calculate the recall at different position (Recall@1, Recall@2, Recall@5) to evaluate the selection of key image frames. Accurate key frame selection indicates precise grounding of text instructions and high executability of the summarization.  %
\subsection{Evaluation Results}
\paragraph{Video Segmentation} 
Table \ref{tab:seg} shows the video segmentation results of different methods. The results indicate that none of the baseline models perform satisfactorily on our dataset. The text-based method outperforms both the vision-based and multi-modal methods. This suggests that existing methods struggle to effectively utilize visual signals of the UIs for video segmentation.

\paragraph{Text Summarization}
Table \ref{tab:textsum} demonstrates the text summarization results of baseline methods. From the results we can observe that baseline methods exhibit substandard performance in term of the Rouge scores. In contrast to segmentation, segmentation, multi-modal methods outperform text-only methods on text summarization, which indicates that the vision signals of the UI layouts can benefit the understanding of text instructions.

\paragraph{Video Summarization}
Table \ref{tab:videosum} shows the video summarization results of baseline methods. The results indicate that the performance of baseline methods is suboptimal, especially regarding the Recall @ 1 metric. This might be attributed to the vision module of baseline models struggling to distinguish similar images of the same UI and detect key operations.

Overall, all baseline methods show unsatisfactory performance on the three core tasks of the proposed dataset, which indicates that the proposed dataset and tasks are challenging and requires specific design for understanding structured and fine-grained UI layouts and actions.

\section{Conclusion}
\label{sec:conclusion}
In this paper, we collect a new video summarization dataset. We firstly focus on UI instructional videos and propose three core tasks for this domain. Experiments demonstrate that existing methods for video summarization struggle with the proposed tasks on the dataset, highlighting the challenge and necessity of the proposed dataset.

\section{Ethical Considerations}
All videos in our dataset are publicly available. We have asked the human workers to make sure that there is no biased and discriminatory content in these videos. For personal privacy, we have anonymized and de-identified any personal information in the videos in the proposed dataset. 
\section{Limitations}
The proposed dataset currently consists solely of UI videos for Adobe Creative Cloud products. We plan to broaden its scope in the future to include a more diverse range of UI videos.

According to the experiments, existing video summarization methods show unsatisfoctory performance on the proposed dataset. It is of vital significance to develop methods which consider the unique features of UI videos to achieve better summarization performance.

\bibliography{custom}
\appendix

\section{Appendix}
\label{sec:appendix}
\subsection{Examples}
In this subsection, we provide several examples of the video summarization data in our dataset. The whole videos and annotations are in the uploaded supplementary data. 

\subsection{Implementation Details}
For Cross Textseg we choose the pre-trained BERT-large model to encode the text folowing the settings of the original paper. For LGSS we follow the implementation in the open-sourced code \footnote{https://github.com/AnyiRao/SceneSeg}. For PySceneDetect we utilize the open-sourced toolkit \footnote{https://www.scenedetect.com}.

For language models, we utilize BERT-large to implement BERT2BERT and utilize T-5-large, BART-CNN-large, BART-XSUM-large model with the Huggingface \footnote{https://huggingface.co/docs/transformers/en/index} toolkit. For MLASK \footnote{https://github.com/ufal/MLASK} and A2Summ \footnote{https://github.com/boheumd/A2Summ} we follow the implementation in the open-sourced code. 

The dataset is split into 8:1:1 for train, validation, and test during training and evaluation.
\begin{figure*}[!t]
\centering
\includegraphics[width=\linewidth]{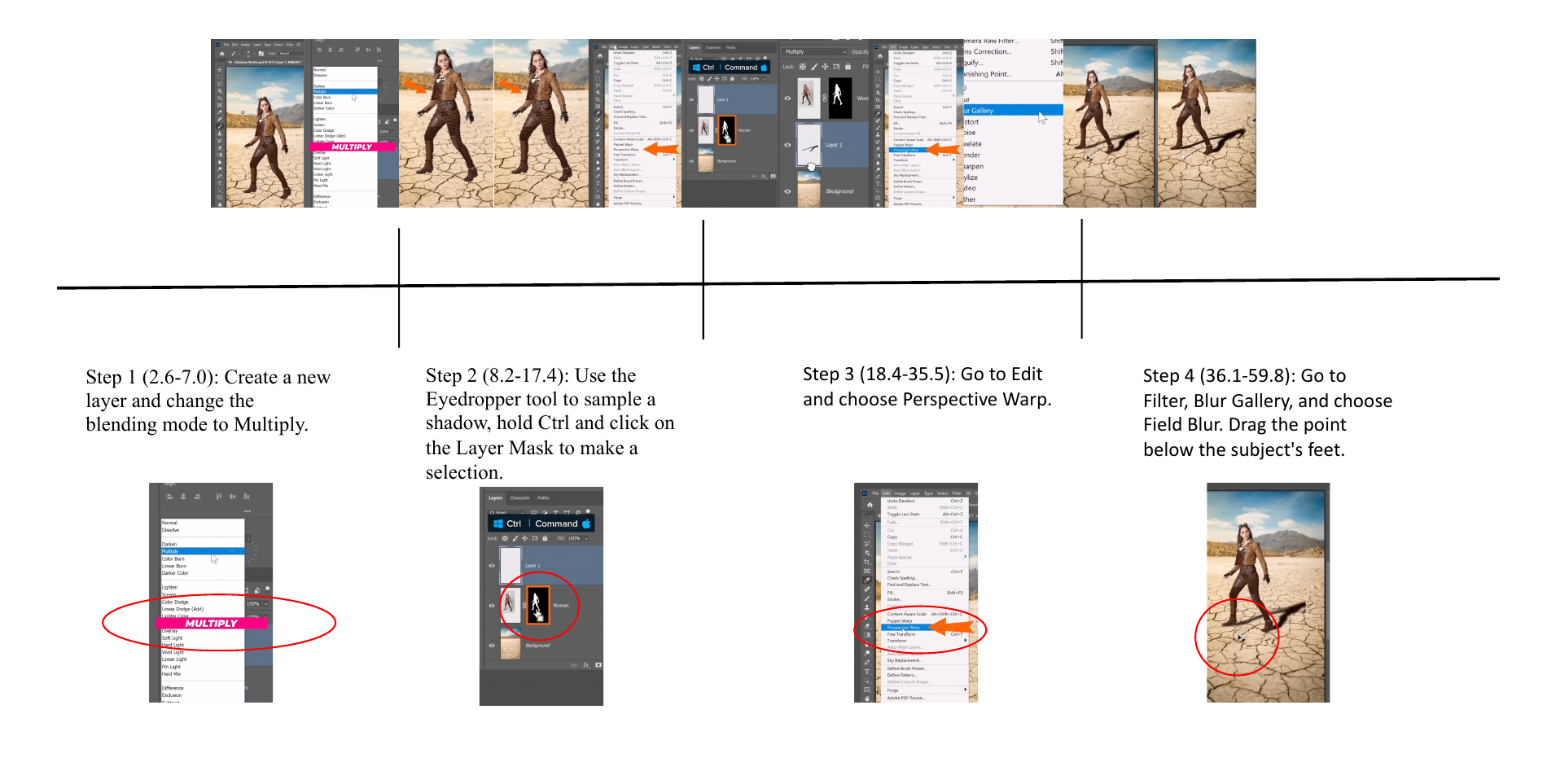}
\caption{An example of the summarization of UI tutorial videos in our dataset. It shows the summarization of the task ``Manipulation of shadows in an image''. First, we segment the video into different steps. For each step, we summarize the key operations into concise text instruction and select a key image frame that describing the corresponding operation.}
\label{fig:example}
\vspace{-0.4cm}
\end{figure*}

\begin{figure*}[!t]
\centering
\includegraphics[width=\linewidth]{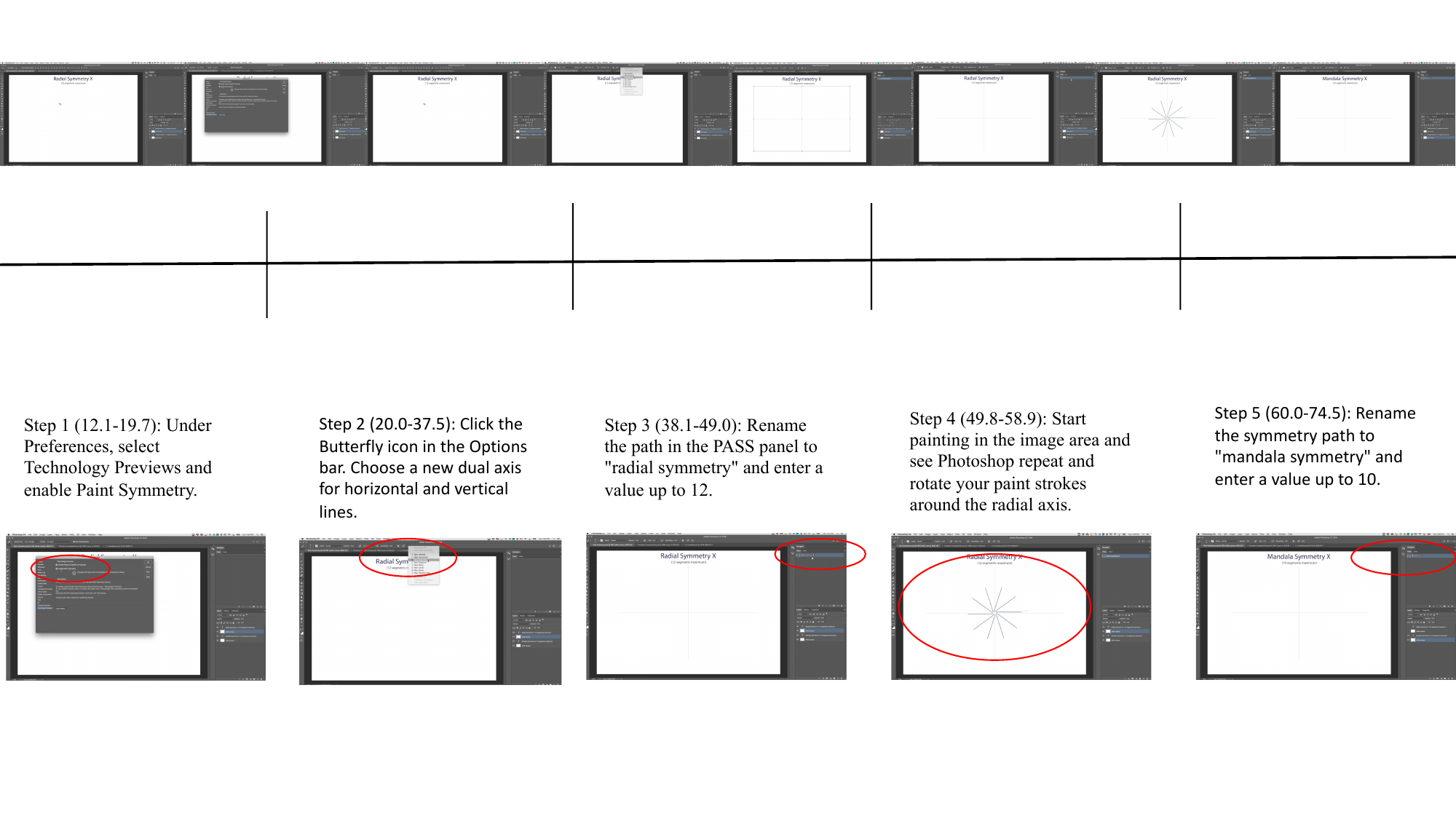}
\caption{An example of the summarization of UI tutorial videos in our dataset. It shows the summarization of the task ``Hidden Features for Using Paint Symmetry''. First, we segment the video into different steps. For each step, we summarize the key operations into concise text instruction and select a key image frame that describing the corresponding operation.}
\label{fig:example}
\vspace{-0.4cm}
\end{figure*}
\end{document}